\documentclass[letterpaper, 10pt, conference]{ieeeconf}
\IEEEoverridecommandlockouts
\usepackage[utf8]{inputenc}

\usepackage{xcolor} 

\usepackage{balance} 
\usepackage{amsmath}
\usepackage{amsfonts}
\usepackage[ruled,vlined,linesnumbered]{algorithm2e}
\SetAlFnt{\small}

\usepackage{adjustbox}
\newcommand\cbox[1]{\raisebox{0.1cm}{\colorbox{#1}{}}}

\usepackage{booktabs}
\usepackage{array}

\newcolumntype{P}[1]{>{\centering\arraybackslash}p{#1}}

\usepackage{graphicx}
\usepackage[font=small]{caption}
\usepackage{subcaption}

\usepackage[
activate   = {true},
protrusion = false,
expansion  = true,
kerning    = true,
spacing    = true,
tracking   = false,
auto       = true,
selected   = true,
factor     = 1000,
stretch    = 10,
shrink     = 10,
]{microtype}

\makeatletter

\renewcommand\paragraph{\@startsection{paragraph}{4}{\z@}%
                                    {0.2em}%
                                    {0em}%
                                    {\hspace{-0.5em}\noindent\bfseries\normalsize}}
\makeatother

\makeatletter
\let\NAT@parse\undefined
\makeatother
\definecolor{hrefcolor}{HTML}{2c778f}
\usepackage[
pdfa,
colorlinks,
bookmarksopen,
bookmarksnumbered,
allcolors=hrefcolor
]{hyperref}
\usepackage[english]{babel}

\usepackage[nameinlink,capitalise]{cleveref}
\crefname{line}{line}{lines}
\crefname{figure}{Fig.}{Figs.}
\Crefname{figure}{Fig.}{Figs.}
\crefname{equation}{Eq.}{Eqs.}
\Crefname{equation}{Eq.}{Eqs.}
\crefname{section}{Sec.}{Secs.}
\Crefname{section}{Sec.}{Secs.}
\crefname{definition}{Def.}{Defs.}
\Crefname{definition}{Def.}{Defs.}
\crefname{algorithm}{Alg.}{Algs.}
\Crefname{algorithm}{Alg.}{Algs.}
\crefname{table}{Tbl.}{Tbls.}
\Crefname{table}{Tbl.}{Tbls.}

\newcommand{\mf}[1]{\mbox{\cref{#1}}\xspace}

\def\xfrom{x_{\text{from}}}
\def\xto{x_{\text{to}}}
\def\xout{x_{\text{out}}}
\def\xstart{x_{\text{start}}}
\def\Xgoal{X_{\text{goal}}}
\def\xrand{x_{\text{rand}}}
\def\xnew{x_{\text{new}}}
\def\xnear{x_{\text{near}}}
\def\bestcost{\textsc{BestCost}}
\def\puzzlealgo{\text{LA-RRT}\xspace}

\def\Esame{E_{\text{same}}}
\def\Ediff{E_{\text{diff}}}

\def\Pbest{P_{\text{best}}}
\def\Ptemp{P_{\text{tmp}}}
\def\ptc{\textsc{ptc}}

\def\dimension{N}
\def\tindex{m}
\def\factor{F}
\newcommand\caction{c_{\text{actions}}}
\newcommand\cactionadd{c_{\text{additive}}}
\newcommand\cdist{c_{\text{dist}}}

\def\Npos{\mathbb{N}_{\geq 0}}

\def\costold{\text{cost}_{\text{old}}}
\def\estart{e_{\text{start}}}
\def\eend{e_{\text{end}}}

\author{Servet B. Bayraktar$^{1}$, Andreas Orthey$^{1, 3}$, Zachary Kingston$^{2}$, Marc Toussaint$^{1}$, Lydia E. Kavraki$^{2}$%
\thanks{This research has been supported by the German Research
Foundation (DFG) under Germany's Excellence Strategy –
EXC 2002/1–390523135 "Science of Intelligence". Research at Rice University has been supported by NSF 2008720.}%
\thanks{$^{1}$TU Berlin, Germany}%
\thanks{$^{2}$Rice University, Houston, TX, USA}%
\thanks{$^{3}$Realtime Robotics, Boston, MA, USA}%
}
\title{\Huge Solving Mechanical Manipulation Puzzles using Asymptotically-Optimal Motion Planning}
\title{\Huge Solving Mechanical Puzzles Optimally using Minimal Action Sequences}
\title{\Huge Solving Logic-Geometric Puzzles Optimally \\using Group-Based Motion Planning}
\title{\Huge Solving Rearrangement Puzzles Optimally \\using Subspace-Decomposition Motion Planning}
\title{\Huge Asymptotically-Optimal Rearrangement Puzzle Planning using Path-Defragmentation}
\title{\Huge Solving Rearrangement Puzzles Optimally\\using Factored Motion Planning}
\title{\huge Solving Rearrangement Puzzles Optimally\\using Path Defragmentation in Factored State Spaces}
\title{\fontsize{22}{26}\selectfont Solving Rearrangement Puzzles\\using Path Defragmentation in Factored State Spaces}

\begin{document}

\maketitle
\begin{abstract}
  Rearrangement puzzles are variations of rearrangement problems in which the elements of a problem are potentially logically linked together. To efficiently solve such puzzles, we develop a motion planning approach based on a new state space that is logically \emph{factored}, integrating the capabilities of the robot through factors of simultaneously manipulatable joints of an object. Based on this factored state space, we propose less-actions RRT (LA-RRT), a planner which optimizes for a low number of actions to solve a puzzle. At the core of our approach lies a new path defragmentation method, which rearranges and optimizes consecutive edges to minimize action cost. 
  We solve six rearrangement scenarios with a Fetch robot, involving planar table puzzles and an escape room scenario. LA-RRT significantly outperforms the next best asymptotically-optimal planner by 4.01 to 6.58 times improvement in final action cost. 
\end{abstract}

\section{Introduction}


%
Rearrangement puzzles are difficult instances of rearrangement problems~\cite{krontiris2015dealing, toussaint2018RSS}, where objects are potentially logically linked to each other. Logically linked objects require the robot to first move one object, before a second can be moved. We define problems with this characteristic as rearrangement puzzles. Those problems are at the core of many robotics tasks like household chores or product assembly. For instance, a robot working in a production facility may need to rearrange objects on a production line, or a robot trapped inside a room has to find its way out (\mf{fig:pullfigure}). 

Rearrangement puzzles are often solved using one of three approaches. In a top-down approach, symbolic reasoning is used to guide exploration of the space~\cite{toussaint2018RSS, Hartmann2021TRO}. 
Such methods include most Task and Motion Planning (TAMP) solvers. TAMP solvers often start by computing action skeletons~\cite{garrett2021integrated}, which are then used to initialize lower-level motion planning~\cite{Grothe2022} or optimization methods~\cite{toussaint2015IJCAI}.
In a bottom-up approach, a search is conducted directly in the joint robot and object state space by carefully analyzing and sampling the constraints involved~\cite{mirabel2018handling, vega2020asymptotically}. 
A third class of methods uses integrated solvers, to selectively switch between lower and higher level abstractions. This can be achieved in different ways, for example using backtracking on failure~\cite{kaelbling2010hierarchical}, or by switching between joint-space sampling and sampling in region where constraint switches occur~\cite{simeon2004manipulation, thomason2022task}. 
Methods for rearrangement puzzles, however, often do not scale well, because they might require excessive backtracking~\cite{20-driess-RSS} or lack good heuristics to guide a solver to a solution~\cite{thomason2022task}. 

To tackle this issue, we propose a complementary method to compute an efficient lower bound on the solution, i.e. an \emph{admissible heuristic}~\cite{koga1994multi, Orthey2020IJRR}. 
This admissible heuristic is obtained by computing feasible paths for the objects alone, while ignoring the robot. 
We show that this admissible heuristic can be efficiently computed, and is able to act as a guide to solve the complete problem involving multiple manipulation actions of the robot. 
\begin{figure}
\centering
\includegraphics[width=\linewidth]{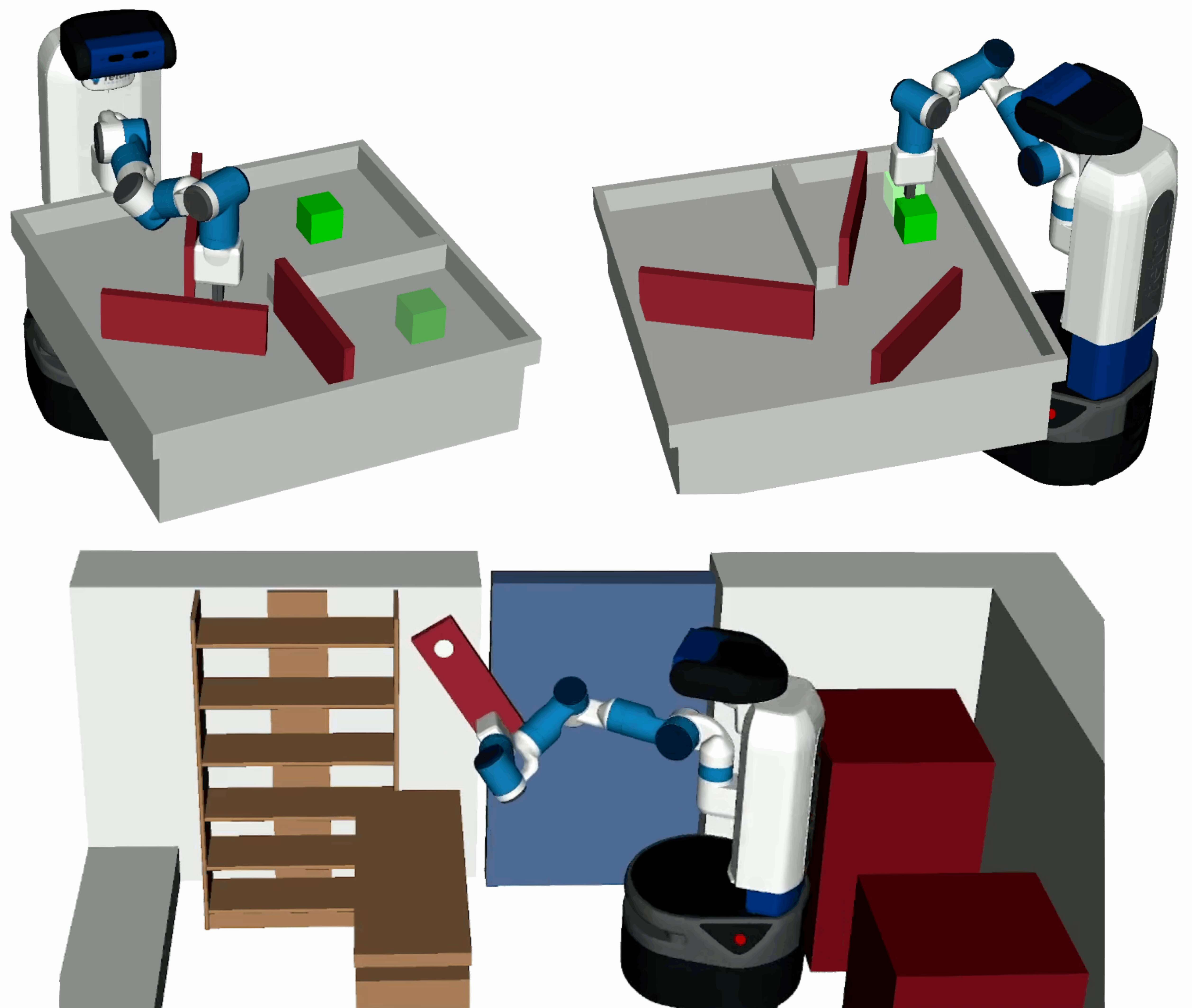}
\caption{Example of a Fetch robot solving a puzzle with 3 doors, to move a green cube from a start to a goal configuration (Top), and a locked room scenario, where the Fetch robot has to escape by rearranging furniture and locks (Bottom).\label{fig:pullfigure}}
\end{figure}

However, computing such an admissible heuristic based on objects alone requires solving two problems. 
The first is the \emph{capabilities problem}. 
If we ignore the robot and its capabilities, solution paths might not be executable by the robot, especially if multiple objects move at the same time---a one-armed robot might be incapable of manipulating two doors at the same time. 
The second is the \emph{minimal actions problem}. 
Computing solutions for the objects alone might produce excessive pick-place sequences, which would be tedious and take too much time for the robot to execute.
\begin{figure*}[t]
    \centering
    \includegraphics[width=\textwidth]{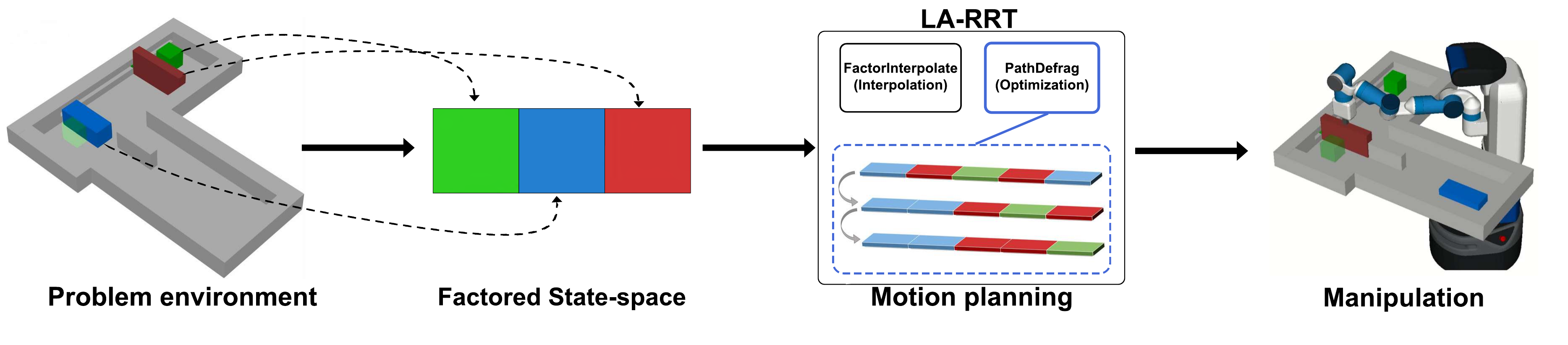}
    \caption{Overview about the system.
    Given a problem environment (\textbf{Left}), we model the objects through a factored state space (\textbf{Middle left}) where each object represents a factor in a different color.
We then apply our algorithm \puzzlealgo to the factored state space, whereby we first exploit the factored interpolation, and then use a path defragmentation method to optimize the number of switches between factors in the resulting path (\textbf{Middle right}). 
This solution is then used to initialize a manipulator algorithm, which computes a complete manipulation sequence to solve the rearrangement puzzle (\textbf{Right}).
    \label{fig:system}}
\end{figure*}

We propose two contributions to address these two issues. 
First, we propose a factored state space to solve the capabilities problem. 
This factored state space implicitly models the capabilities of the robot by grouping joints into \emph{factors}, wherein joints in a factor are simultaneously manipulatable by a given robot. 
We develop novel interpolation and action cost functions to make this state space usable by general-purpose motion planners~\cite{Lavalle2006, karaman2011sampling}. 
Second, we propose a new method to significantly reduce the number of actions, which we call \emph{path defragmentation}. 
Path defragmentation can reduce the number of actions by reasoning about the ordering of path segments through different factor spaces. 
This method is integrated into a new planner, the less-actions RRT (\puzzlealgo). 
\puzzlealgo is able to handle the non-additive property of our minimal-actions cost, and can split interpolated path segments into sequences of factored path segments, which are then optimized using path defragmentation.
We apply \puzzlealgo to difficult instances of rearrangement puzzles where we assume that the mode of each object is constant and show that it can find paths with significantly less action cost compared to state-of-the-art planners. 
Eventually, we use those paths to compute complete manipulation sequences for a simulated Fetch robot. Fig.~\ref{fig:system} shows an overview about our method.

\section{Related Work}

We group approaches to rearrangement problems into three groups, namely top-down planning, bottom-up planning, and integrated planning approaches.

Top-down approaches are often based on computing \emph{action skeletons}, sequences of symbolic actions which are then used as constraints for joint-space planning or optimization approaches~\cite{toussaint2015IJCAI, kaelbling2010hierarchical}. This approach is highly successful if the degrees of freedom in the environment are logically decoupled, like pick-place actions on distinct objects~\cite{krontiris2015dealing, garrett2018ffrob, garrett2018sampling}. It is often sufficient in these scenarios to compute joint-space trajectories without feedback, even if multiple robots and large planning horizons are involved~\cite{toussaint2018RSS, Hartmann2021TRO}. However, if an action skeleton cannot be solved expensive backtracking is required~\cite{20-driess-RSS} to validate another action skeleton. 

Our work is complementary, in that we also execute symbolic actions on our robot, but we choose those actions \emph{implicitly} using the (admissible) heuristic from our \puzzlealgo planner, which gives us a complete valid sequential solution path for all objects involved. This tighter integration of planning and robot capabilities give us a better chance to avoid expensive backtracking.

Another approach to rearrangement planning are bottom-up approaches. These approaches explicitly search through the combined space of the robot and objects~\cite{thomason2019unified}. 
This can be advantageous as sampling-based planners can be applied~\cite{kingston2018sampling}, which can achieve asymptotic optimality guarantees~\cite{vega2020asymptotically, shome2020pushing, schmitt2017optimal}. 
These methods often use a constraint-graph~\cite{mirabel2017manipulation, mirabel2018handling}, which enumerates the valid constraint combinations in a scene. 
Constraint-based methods require effective projection methods~\cite{berenson2011task, kingston2018sampling} to sample and interpolate along constrained subspaces. 
While these approaches can provide strong guarantees~\cite{vega2020asymptotically}, scaling to higher dimensions (i.e., number of objects) adds significant challenges~\cite{kingston2022tro}.

Our approach differs in that we do not plan directly in the full robot-object configuration space, which can be costly to compute. Instead, we plan first in the object-only configuration space by integrating the capabilities of the robot into the motion planner using the new factored state space (see \mf{sec:problemstatement}).

Finally, rearrangement problems can be solved with integrated methods. 
Those methods combine both planning on a high-level, like a symbolic layer, with planning on a lower level, like joint-space planning. 
A tight integration is crucial to avoid backtracking on infeasible high-level solutions. Hierarchical planners~\cite{barry2010hierarch,kaelbling2010hierarchical} can often find good solutions with few backtracking operations. 
However, when objects are logically coupled, like in Navigation among moveable obstacles (NAMO) problems~\cite{stilman2008planning}, it is often difficult to find the correct high-level solution. 
To tackle this issue, geometric information can often be integrated into the symbolic description~\cite{garrett2018ffrob}, or sampling is extended to include joint configurations consistent with symbolic actions~\cite{simeon2004manipulation, thomason2022task}. 
While those approaches provide concise frameworks for optimal planning, they might suffer from slow convergence due to the high branching factor of possible actions. 

To tackle the high branching factor, it becomes often necessary to find good heuristics. 
Recent approaches include reasoning about collision regions between objects \cite{wang2021uniform}, or by computing initial solutions, where every object is moved at most once (the monotone case)~\cite{wang2022efficient}. 
Our approach is complementary by also computing a heuristic. However, our heuristic is admissible, and is tailored to problems where objects are logically linked to each other. 

The computation of this admissible heuristic is done using the new LA-RRT planner, which can efficiently reason over factored state spaces. This planner is inspired by planners like the Manhattan-like RRT (ML-RRT)~\cite{Cortes2007}. In ML-RRT, planning is decomposed into an active and a passive subspace. Our planner LA-RRT, however, generalizes this idea to arbitrary subspaces (factors), makes it applicable to manipulation, and combines it with optimality, such that we approach the minimal number of switches between factor spaces.
\section{Factored State Space\label{sec:problemstatement}}

Our goal is to solve manipulation planning problems that require solutions through the combined state space of $X_R \times X$, where $X_R$ is the robot's state space and $X$ is the state space of the manipulatable objects. We decompose this problem by projecting this state space $X_R \times X \rightarrow X$ removing the robot state space. 
To account for the robot capabilities, we manually define instead a factored decomposition of $X$ as $X= F_{1} \times F_{2} \times \ldots \times F_{N} $ as shown in \mf{fig:fragmentstatespace}, where a factor represents a subspace of $X$ containing joints, which are \emph{simultaneously manipulatable} by the robot. For convenience, we also define the operation \textit{$k$ in $F_i$} as returning the indices of the joints in $F_i$, and the operation $|F_i|$ as the number of joints.
\paragraph{Example} For a one-arm manipulator arm, a door with
a single revolute joint (depicted as $F_{2}$ in red in \mf{fig:fragmentstatespace}), would have $|F_{door}|$ = 1, while a
movable planar disc would have $|F_{disc}|$ = 2 (shown as $F_{1}$ in green in \mf{fig:fragmentstatespace}).

The purpose of the factors is to ensure that at most $k$ objects move at the same time given $k$ available manipulator arms. This constraint is an implicit way to define \emph{action skeletons}~\cite{Hartmann2021TRO}, i.e., as a sequence of alternating factor-paths, which can then be executed individually by executing pick-place actions with the robot. 

\begin{figure}
\centering
\includegraphics[width=\linewidth]{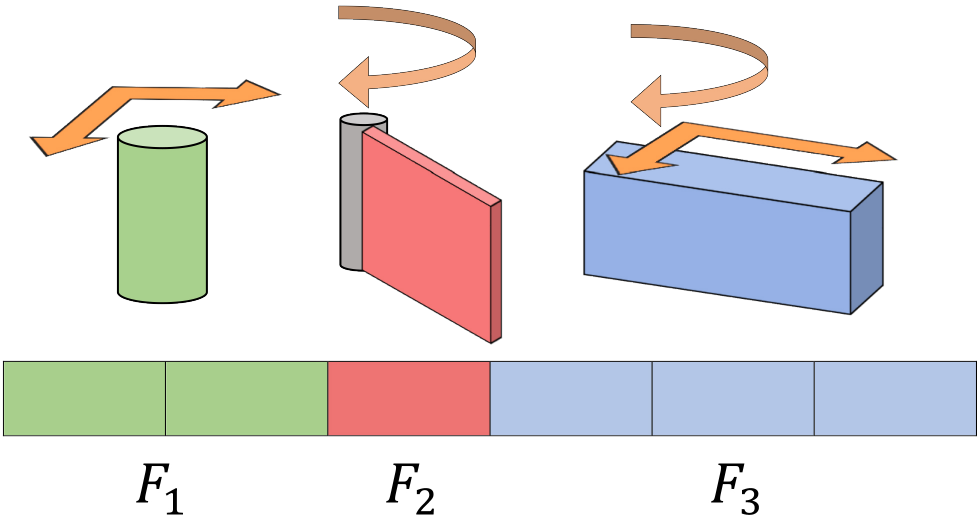}
\caption{Example of the structure of the factors in the state-space. Each color represents a factor, and each joint of an object corresponds to a cell in the factor. \label{fig:fragmentstatespace}}
\end{figure}

To plan in factored state spaces requires the implementation of three functionalities which are crucial for planning: interpolation, goal constraint, and cost.

\paragraph{Interpolation} A linear interpolation between two states would interpolate in all dimensions thereby moving objects simultaneously. 
To only move one factor at a time, we develop a Manhattan-like interpolation method applicable with arbitrary factor state spaces, which is depicted in \mf{alg:groupinterpolate}. 
This method interpolates a path between two states $\xfrom$ and $\xto$ by using an interpolation parameter $t \in [0,1]$, and an ordering of factor spaces\footnote{In our evaluations, we use a random ordering, since we found no significant influence on performance for different orderings.}. 
As output, we return a state $\xout$ at distance $t$ between $\xfrom$ and $\xto$. 
This methods works by first computing the individual distances between $\xfrom$ and $\xto$ when \emph{projected} onto each factor (Line 1). 
We then find the factor space $\tindex$ where the interpolation variable $t$ lies (Line 2). 
As an example, let us assume we have two factors with distances $1$ and $3$, and total distance $4$. 
In the first step, we normalize them to $\frac{1}{4}$ and $\frac{3}{4}$. If $t<\frac{1}{4}$, we select the first factor. 
If $t \geq \frac{1}{4}$, we select the second factor.

All factor spaces before $\tindex$ are fully interpolated and set to the corresponding values of $\xto$ (Line 4--6). Then, we call the intrinsic interpolation function of the selected factor $F_{\tindex}$ and change its corresponding indices in $\xout$ (Line 8). Finally, we set the factors after $\tindex$ to the $\xfrom$ values (Line 9--10), and return $\xout$ (Line 11).
\begin{algorithm}
\DontPrintSemicolon
\KwIn{$\xfrom$, $\xto$, $t$, ${\factor_1,\ldots,\factor_M}$}
\KwOut{$\xout$}

\SetKwFunction{getDistances}{GetDistances}

$d_{1:M} \longleftarrow \ \textsc{GetDistances}(\xfrom,\xto,\factor_{1:M} )$\;
$\tindex \longleftarrow \textsc{findIndex}(d_{1:M}, t)$\;
$d_{interpolated} \longleftarrow 0$\;
\For{$i=1$  \KwTo  $\tindex-1$}{
    \lForEach{$j$ \textnormal{\textbf{in}} $\factor_{i}$}{
        $\xout[j] \longleftarrow \xto[j]$
    }
    $d_{interpolated} \mathrel{+}= d_i$\;
}

$s \longleftarrow $ $(t-d_{interpolated} / d_\tindex)$ \;

$\xout \longleftarrow \textsc{Interpolate}(\factor_{\tindex}, \xfrom, \xto, s ) $ \;


\For{$i=\tindex+1$ \KwTo $\dimension$}{
    \lForEach{$j$ in $\factor_{i}$}{
        $\xout[j] \longleftarrow \xfrom[j]$
    }
}
\Return $\xout$\;

\caption{\textsc{FactorInterpolate}\label{alg:groupinterpolate}}
\end{algorithm}

\paragraph{Goal} Next, we define a goal constraint by taking all joints into account which have a designated goal position. We call this set the goal-indices $G_{I_G}$. To take joint positions into account, we use a goal state $x^g$ which is only defined for indices in $G_{I_G}$. All other indices are freely-chooseable and can be randomly sampled. For such a partial state $x^g$ we represent an \emph{$\epsilon$-goal region} $X_G$ as
\begin{equation}
    X_G = \{x \in X\mid d(x_i, x^g_i) \leq \epsilon, i \in G_{I_G}\}.
\end{equation}
%

\paragraph{Cost} Finally, we have to define a cost function, which reflects our desire to minimize the number of pick-place actions the robot has to execute. 
This can be done by counting the number of factor changes when interpolating between two states, i.e., $\caction: X \times X \rightarrow \Npos$. 
However, this cost is difficult to define as we show in \mf{fig:cost-explanation}, because of two issues. First, we cannot discriminate between paths of different lengths ($p_1$ and $p_3$ in \mf{fig:cost-explanation}). 
We resolve this by adding a multi-layered cost function which works as $\caction$, but behaves like a distance cost $\cdist$ when the number of actions are equivalent.
The second issue is related to the additive nature of this cost function. Planners like BIT* and RRT* assume that cost terms can be summed along path segments~\cite{karaman2011sampling}. 
However, such an additive cost function $\cactionadd$ cannot discriminate between path segments which have equivalent actions along subsequent edges ($p_1$ versus $p_2$ in \mf{fig:cost-explanation}). 
Planners running with $\cactionadd$ will be able to find the right \emph{equivalence class} of solution paths ($p_1$ or $p_2$), but only by using the non-additive version of $\caction$ can we pick the correct solution path ($p_1$).
This issue forces us to develop a dedicated planning algorithm which can correctly exploit the non-additive nature of $\caction$. 

\begin{figure}
    \centering
    \includegraphics[width=0.9\linewidth]{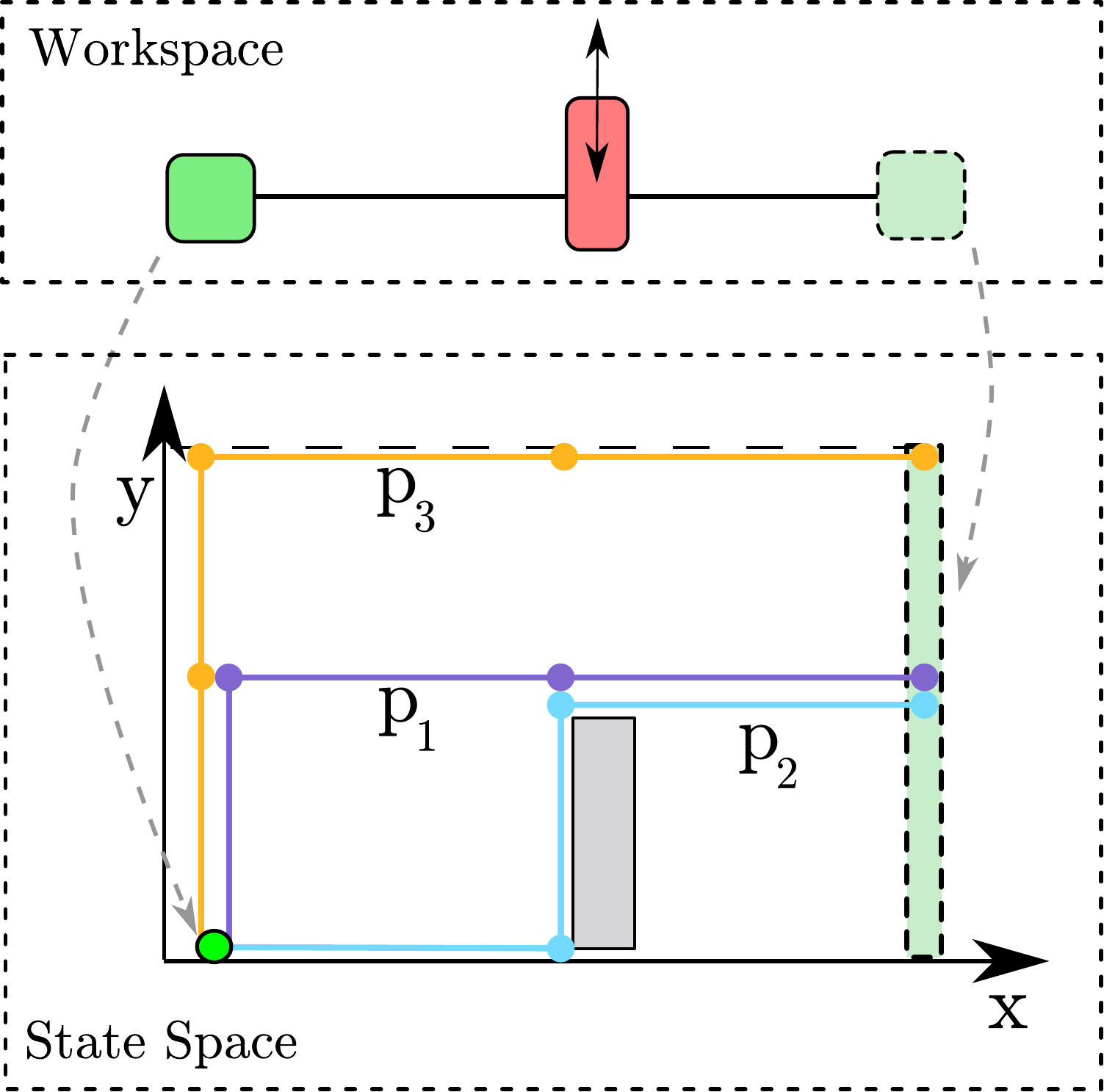}
    \caption{Explanatory example to demonstrate the difference between different cost functions. \textbf{Top}: Workspace of a single cube on a rail, which has to be moved from its start configuration (left) to a desired goal configuration (right). X-axis is the horizontal position of the green cube and the y-axis is the vertical position of the blocking cube. A blocking cube in the middle (red) needs to be moved out of the way to solve the problem. \textbf{Bottom}: The state space of this problem with the start configuration (green), goal region (lightgreen) and the collision-region (gray). We showcase three paths. $p_1$ is the optimal path with cost terms $\caction=2, \cactionadd=3, \cdist=3$, $p_2$ with $\caction=3, \cactionadd=3, \cdist=3$, and $p_3$ with $\caction=2, \cactionadd=4, \cdist=4$. This highlights two issues: (a) To discriminate between $p_1$ and $p_3$, we require a multi-layered cost function taking also distance into account, and (b) planner like BIT* and RRT*, which support only \emph{additive} costs, cannot discriminate between $p_1$ and $p_2$. Our planner LA-RRT explicitly works on \emph{non-additive} costs, and can thereby correctly identify $p_1$ as low cost action.}
    \label{fig:cost-explanation}
\end{figure}

Having defined the factored state space, we can define the problem of solving a rearrangement puzzle as a factored state space $X$ together with a start configuration $x_s$, a goal region $X_G$, and a cost function $\caction$. Our goal is to find a path from $x_s$ to $X_G$ minimizing $\caction$. 
\section{Less-Actions RRT}

Less-actions RRT (\puzzlealgo) is a bi-directional planner modelled after RRT-Connect~\cite{kuffner2000rrt} and RRT*~\cite{karaman2011sampling} to efficiently search over factored state spaces with non-additive action costs. \puzzlealgo differs by using a different extend method and a novel path optimization method, which we call \emph{path defragmentation}.

An overview of LA-RRT is given in \mf{alg:larrt}. As in RRT-Connect~\cite{kuffner2000rrt}, we initialize a start tree $T_{a}$ and keep a set of goal trees $T_{b}$ from at most $M$ sampled states in our goal region (Line 2). We define the variable $\bestcost$ as the best action cost and $\Pbest$ as the best solution path found (Line 3--4). Like in RRT-Connect~\cite{kuffner2000rrt}, we alternate tree expansion by sampling a random motion $\xrand$ (Line 6) and extend the corresponding tree (Line 7). Once the trees can be connected (Line 8), we apply the path defragmentation algorithm to the solution path~(see \mf{sec:pathdefragmentation}).
If the new path has a better cost than the current best cost, we save $\Ptemp$ as the new best path (Line 11--13), and continue searching until we reach the planner termination condition $\ptc$.

\begin{algorithm}
\DontPrintSemicolon

\SetKwRepeat{Do}{do}{while}
\KwIn{$\textsc{ptc}, \xstart, \Xgoal$}
    $T_{a}$.init($\xstart$), $T_{b}$.init($\Xgoal$)\;
    $\bestcost \longleftarrow \infty$\;
    $\Pbest \longleftarrow$ [ ] \;
    \While{$\neg \textsc{ptc}$}
    {
        $\xrand =$ \textsc{sampleConfiguration}()\;
        $ treeInfo \longleftarrow  \textsc{FactorExtend}(T_{a},\xrand)$\;
        \If{$ treeInfo  \neq TRAPPED$}
        {
            \If{$ treeInfo = REACHED$}
            {
                $\Ptemp \longleftarrow$ \textsc{Path}($T_{a},T_{b}$)\;
                
                \textsc{PathDefragmentation}($\Ptemp$)\tcp*{\ref{sec:pathdefragmentation}}
                
                \If{$ \textsc{COST}(\Ptemp) < \bestcost $}
                {
                 $\bestcost =$ \textsc{Cost}($\Ptemp)$\;
                 $\Pbest$ = $\Ptemp$\;
                }
                 
            }
        }
        \textsc{SWAP}($T_{a}$,$T_{b}$)\;
    }
    \Return $\Pbest$\;

\caption{\textsc{LA-RRT}\label{alg:larrt}}
\end{algorithm}

\subsection{Factor extend and splitting edges \label{sec:stateisolation}}

Contrary to RRT-Connect, LA-RRT needs to take the factors into account when extending states. This is accomplished by the \textsc{FactorExtend} method (\mf{alg:extend}). \textsc{FactorExtend} extends a random sample $\xrand$ by doing a factor interpolation to find a new state $\xnew$ (Line 2), and checks if it is valid (Line 3), as in the original RRT-Connect~\cite{kuffner2000rrt}. 

If the edge is valid, we call the \textsc{SplitEdge} method, which splits an edge into a sequence of sub-edges, whereby each sub-edge only changes one factor at a time (Line 4-6). This is visualized in \mf{fig:motioncostfigure}. By splitting the edge, we thus ensure that each sub-edge has minimal action cost.
If the result is collision-free, we add the sub-edges to our tree (Line 5), or return with a failure (Line 6). 
If we successfully added the sub-edges to the tree, we finally return the status of the extension (Line 7-9).

\begin{figure}
\includegraphics[width=\linewidth]{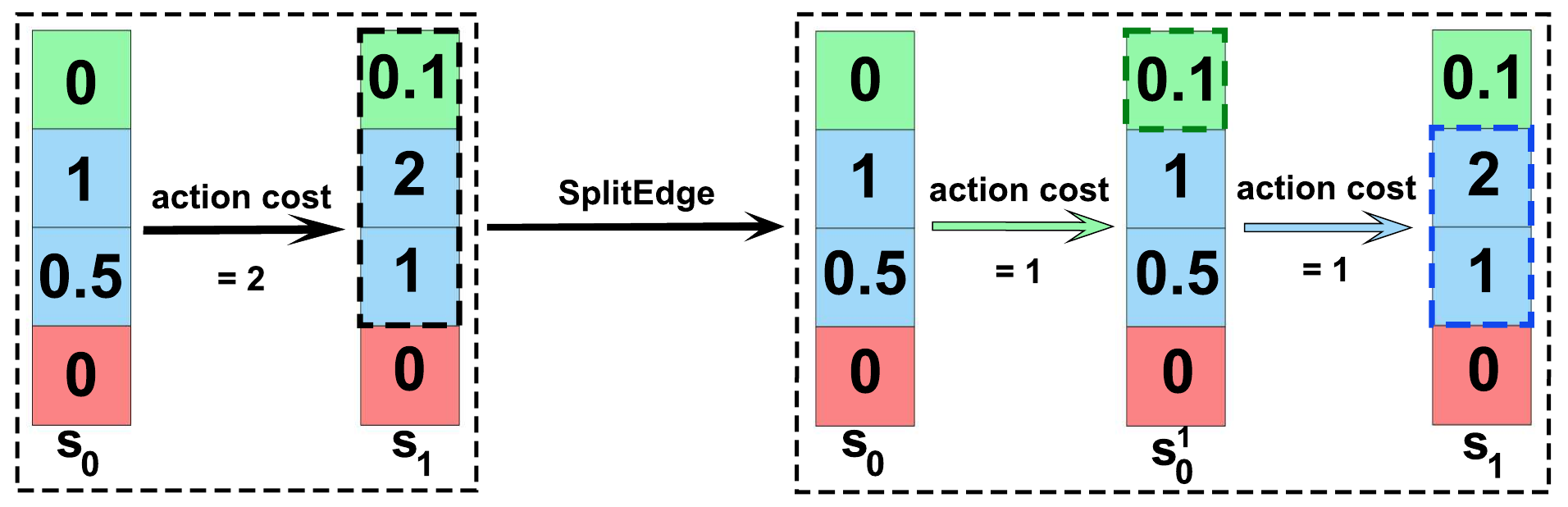}
\caption{Example of the \textsc{SplitEdge} process ensuring minimal action cost. The edge between states $s_0$ and $s_1$ (left box) has an action cost of $2$ changing the green and blue factors. Each color represents a factored state space with values corresponding to object joint values. Applying \textsc{SplitEdge} creates two edges, which changes first the green factor space, then the blue factor space. This results in two states, $s_{0}^1$ and $s_1$. \label{fig:motioncostfigure}}
\end{figure}

\begin{algorithm}
\DontPrintSemicolon
\KwIn{$T$, $\xrand$}

\tcc{Factored extension of $\xrand$} 
    $\xnear \longleftarrow T.nearest(\xrand)$\;
    $\xnew \longleftarrow \textsc{FactorInterpolate}(\xnear,\xrand)$\;

    \lIf{$ \neg \textsc{IsMotionValid}(\xnear,\xnew)$}{ \Return $TRAPPED$}

\tcc{Splitting edge into factors}

            $X_{iso} \longleftarrow \textsc{SplitEdge}(\xnear,\xnew)$\;

            \lIf{\textsc{areStatesValid}($\xnear,X_{iso})$}{
                $T.add(X_{iso})$
            }
            \lElse{ \Return $TRAPPED$}
\tcc{Return status of extension}
    \If{\textsc{distance}($\xnear,\xrand) > maxDistance$}{\Return $ADVANCED$\;}
    \lElse{ \Return $REACHED$}

\caption{\textsc{FactorExtend}\label{alg:extend}}
\end{algorithm}

\subsection{Path defragmentation\label{sec:pathdefragmentation}}

The obtained paths using the factored extend method are often highly fragmented, meaning they exhibit frequent factor switches.
To tackle this issue, we develop the \textsc{PathDefragmentation} method. This method takes as input a path and reduces its number of factor switches. 

This path defragmentation method is summarized in Alg. \ref{alg:pathdefragmentation}. We assume that input paths only contain edges changing at most a single factor at a time. For simplification, we say that each edge has an associated factor, meaning the edge leads to changes in a particular factor space.

We start with an input path $P$ containing multiple factor switches. We obtain all edges in $P$ (Line 3) and set the first edge to $\estart$ (Line 4). We then iterate over all edges until we reach the last edge (Line 5). During the iteration, we first check if the next edge has an equivalent factor, in which case we continue (Line 6-8). 

If the factors of $\estart$ and the next edge mismatch, we identify the next block of edges with factors identical to $\estart$ (Line 9-11). 
First, a forward search (Line 9) finds the edge $\eend$, which is the first edge \emph{after} the next consecutive block of the edges having the same factor as $\estart$ (see Fig.~\mf{fig:pathdefrag}). We then extract this block of edges (Line 10), together with the edges which differ in between (Line 11), storing them in the variables $\Esame$, and $\Ediff$, respectively.
The resulting sets of edges are shown in \mf{fig:pathdefrag}.

Given $\Esame$ and $\Ediff$, we try to reorder them as shown in \mf{fig:pathdefrag}.
To reorder the edges, first we rewire the current edge $\estart$ to the set $\Esame$ and then to the set $\Ediff$ (Line 12). 
The function \textsc{ReorderEdges} also does collision checking during the reordering process and returns a reordered set of edges (a path segment) if all edges from $\Esame$ and $\Ediff$ have been successfully connected. 
If a collision occured or edges failed to get connected, we return an empty path. 
If the reordering is successful (Line 13), the function \textsc{ReplacePathEdges} is called (Line 14) to replace the corresponding edges in the original path with the reordered edges. Finally, we update $\estart$ by setting it to the first edge of $\Ediff$ (Line 15) as shown in the lower part of \mf{fig:pathdefrag}. This process is repeated until we cannot improve the cost (Line 16).

After convergence, we use two methods as post-optimization steps. First, we utilize the \textsc{trySkipFactor} method in which we try to skip factors that are not mandatory to reach the goal state (Line 16). Those are removed from the path. Second, we use the \textsc{simplifyActionIntervals} (Line 17), where we attempt shortcuts between the same edges having the same factor switches (\mf{fig:pathdefrag}).

\begin{figure}
\centering
\includegraphics[width=\linewidth]{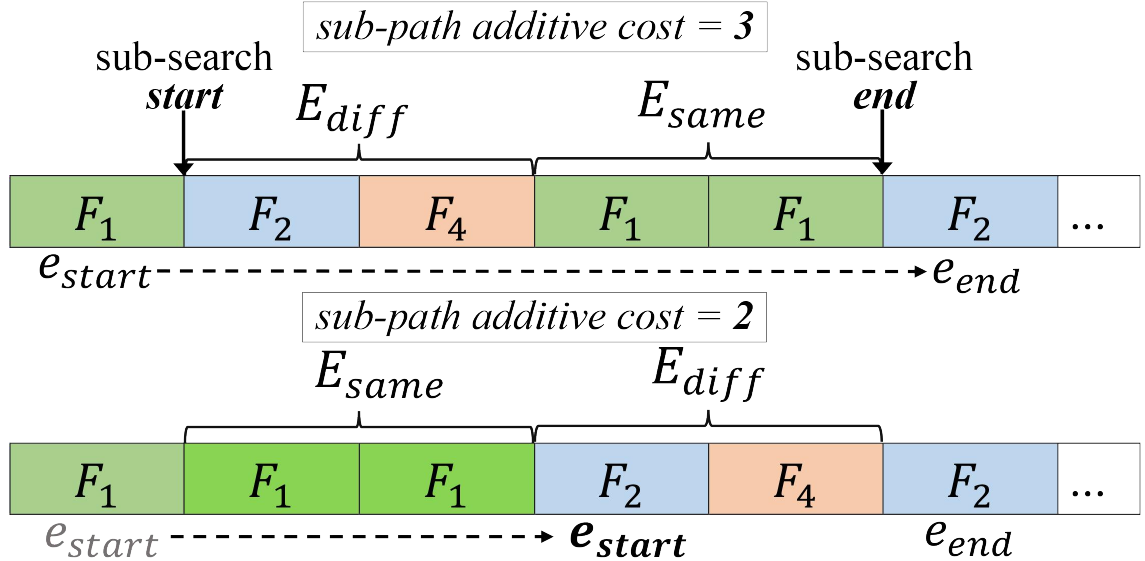}

\caption{Example of one successful rewiring iteration in \textsc{PathDefragmentation} method. Each block represents an edge between two states in which only a single factor is changed, whereby equivalent factor spaces have the same color. The upper path has a cost of $3$ between states $1$ to $5$. After one iteration, the additive cost decreases to 2 in the lower path. \label{fig:pathdefrag}}
\end{figure}
\begin{algorithm}
\DontPrintSemicolon
\KwIn{$P$}
\SetKwRepeat{Do}{do}{while}

\Do{ $\textsc{Cost}(P) < \costold $ }
                {
    $\costold \longleftarrow$ \textsc{Cost}$(P)$ \;

    $E \longleftarrow$ \textsc{GetEdges}($P$)\;
    $\estart \longleftarrow E(0)$\;
    
    \While{\textsc{Next}($\estart$) $\neq$ NULL}{
        
        
        \If{\textsc{EquivalentFactors}$(\estart, \textsc{Next}(\estart))$}
        {
            $\estart \longleftarrow \textsc{Next}(\estart)$\;
            \textsc{Continue}\;
        }
        $\eend \longleftarrow \textsc{FindEndEdge}(\estart)$\;
        $\Esame \longleftarrow$ \textsc{SameFactorEdges}($\estart,\eend$)\;
        $\Ediff \longleftarrow$ \textsc{DiffFactorEdges}($\estart, \eend$)\;
           
        $P' \longleftarrow$ \textsc{ReorderEdges}$(\Esame, \Ediff)$\;
           
        \If{$P'$}{
             \textsc{ReplacePathEdges}$(P, P', \estart, \eend)$\;
         }
         $\estart \longleftarrow \textsc{UpdateStartEdge}(\estart, P')$\;
    }
}
\textsc{trySkipFactor}($P$)\;
\textsc{simplifyActionIntervals}($P$)\;
\caption{\textsc{PathDefragmentation}\label{alg:pathdefragmentation}}
\end{algorithm}

\section{Manipulation with \puzzlealgo}

After \puzzlealgo converges, we use the attained factored paths to manipulate the objects with a Fetch robot as shown in \mf{fig:pullfigure}. 
This requires that we plan pick and place motions of the robot for each segment of the solution path, to actually actuate the objects.
The manipulation of the results consists of four main steps. We first extract the actions from the solution path and we match it with the objects in the environment. Then we compute valid random grasp positions which are simultaneously feasible at the start and goal state of the object. Finally, we create a task-space region~\cite{berenson2011task} which encodes the constraints to transport the object. We then plan for this motion with KPIECE~\cite{csucan2009kinodynamic}. If successful, we advance to the next action, or we try again by grasping a different point on the object. Note that this is just one possible way to exploit paths from \puzzlealgo for manipulation planning. Integrating those paths into more powerful manipulation frameworks~\cite{toussaint2018RSS, garrett2018ffrob} could yield improved results.



\section{Evaluation}

\def\figWidth{0.32\linewidth} 

\def\subfigWidth{0.49\linewidth} 
\def\figWidth{\linewidth} 
\begin{figure}[!tbp]
    \centering
    \begin{subfigure}[b]{\subfigWidth}       \includegraphics[width=\figWidth,keepaspectratio]{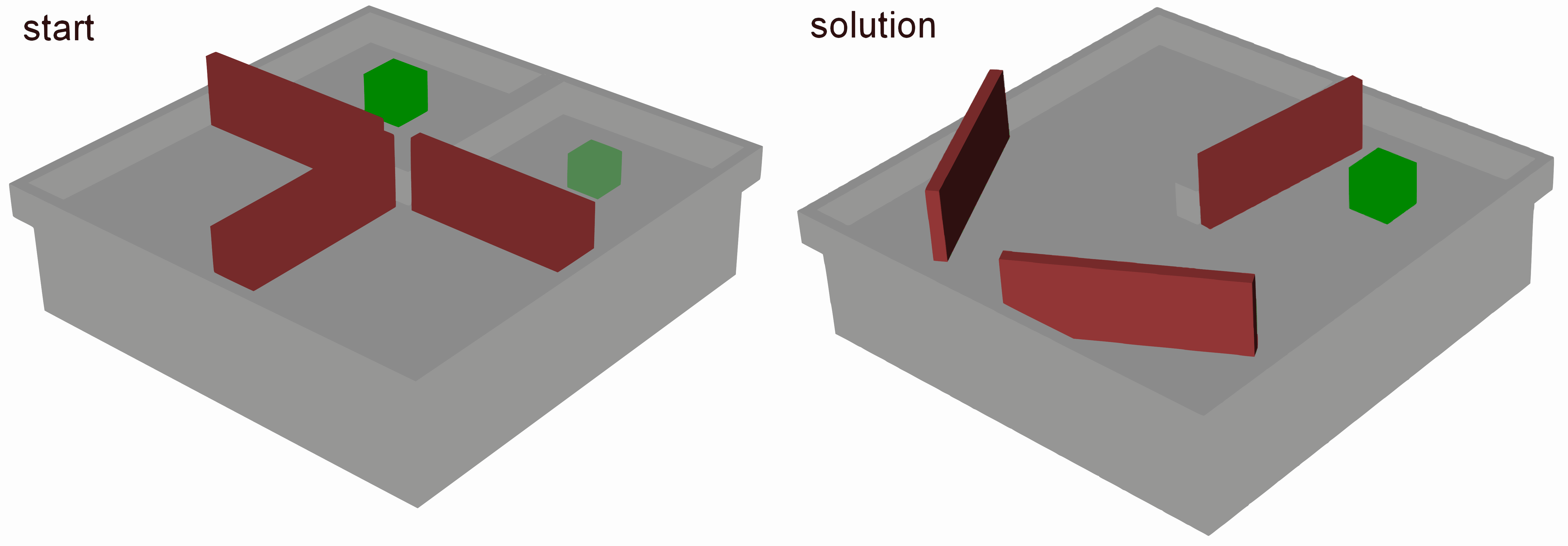}
    \caption{Maze 3 Doors\label{fig:exp:maze3doors}}
    \end{subfigure}
    \hfill
    \begin{subfigure}[b]{\subfigWidth}       \includegraphics[width=\figWidth,keepaspectratio]{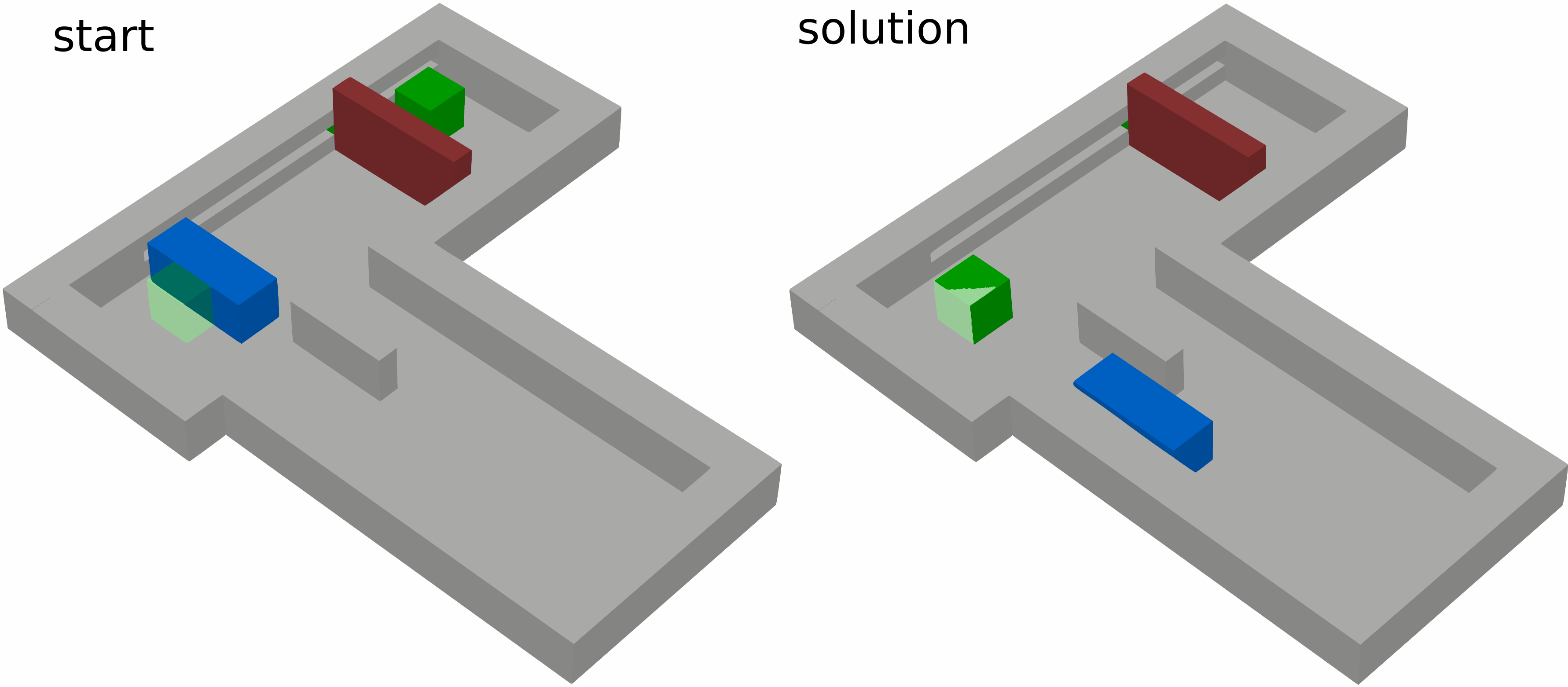}
    \caption{Maze Slider \& Obstacle\label{fig:exp:mazeslider}}
    \end{subfigure}
    
    \begin{subfigure}[b]{\subfigWidth}       \includegraphics[width=\figWidth,keepaspectratio]{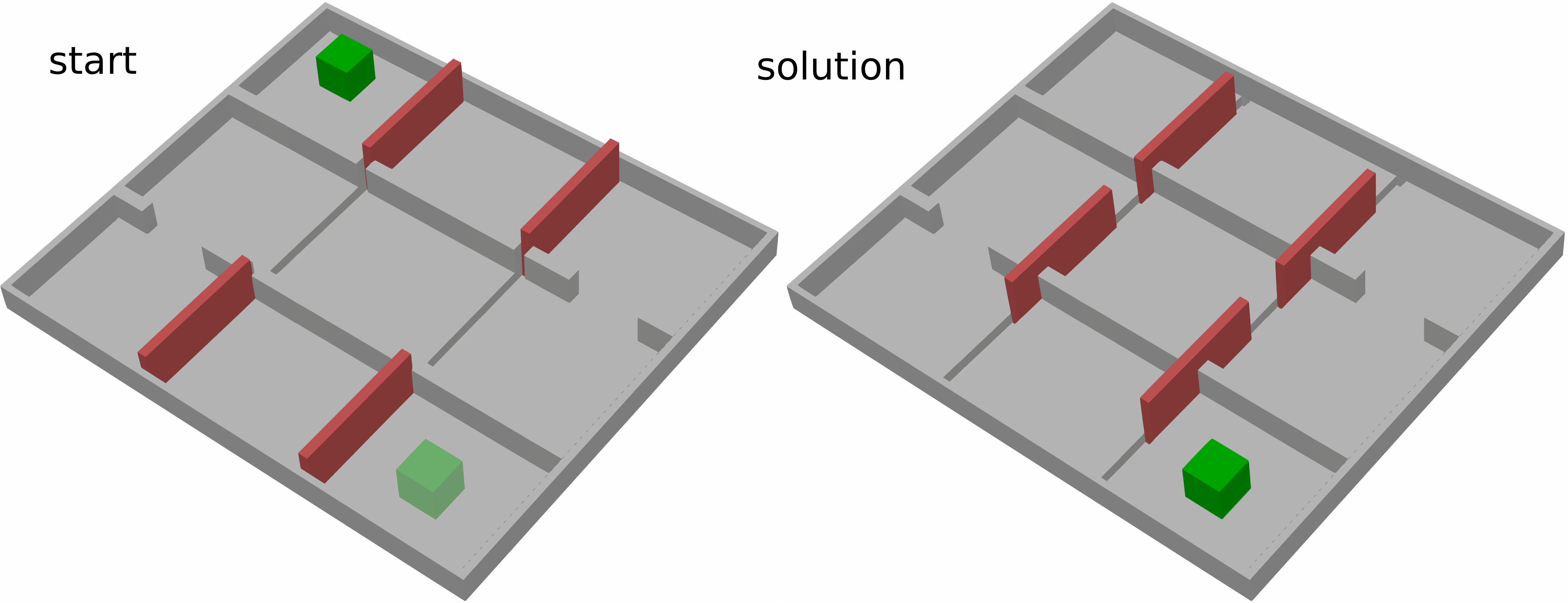}
    \caption{Maze 4 Sliders\label{fig:exp:maze4sliders}}
    \end{subfigure}
    \hfill
    \begin{subfigure}[b]{\subfigWidth}       \includegraphics[width=\figWidth,keepaspectratio]{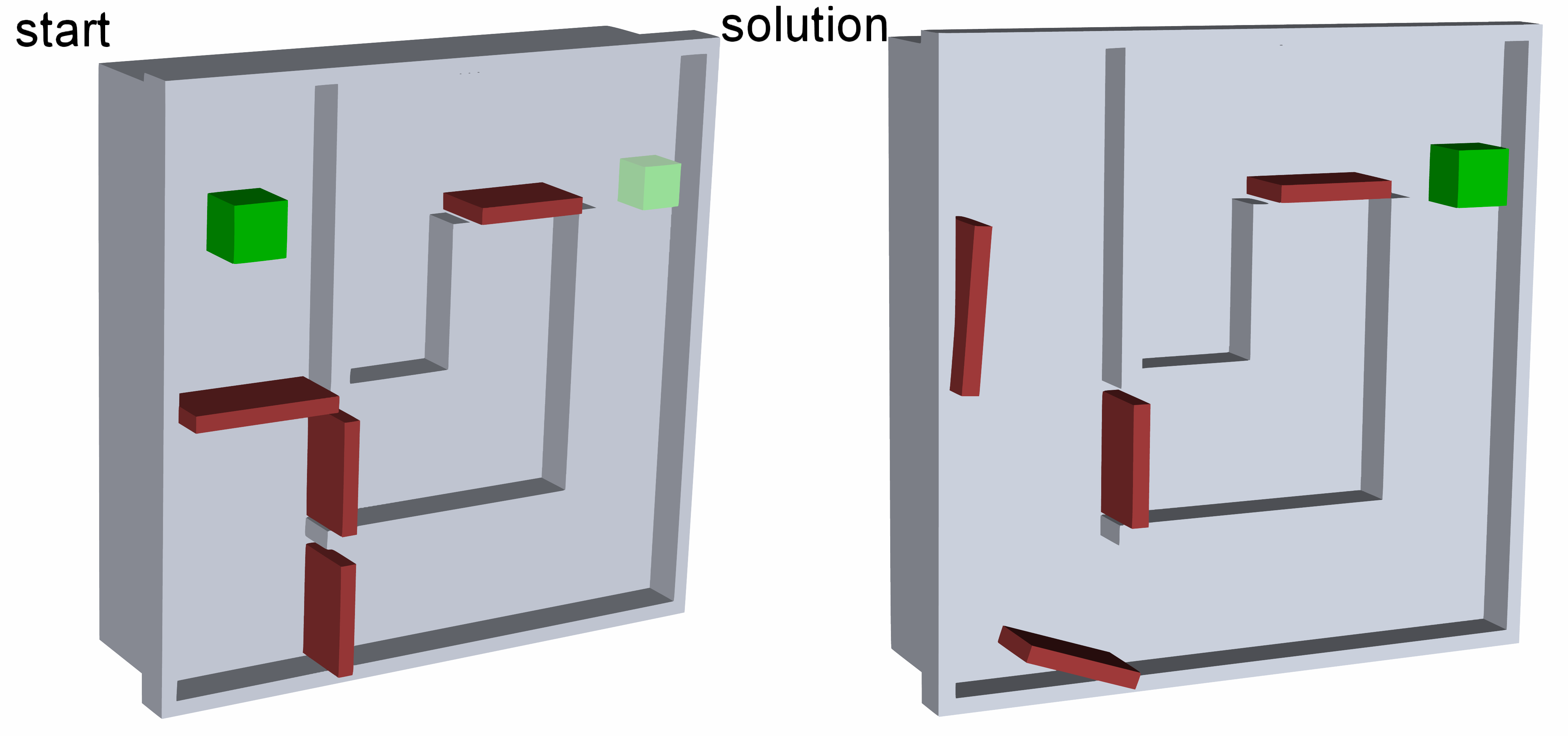}
    \caption{Maze Vertical\label{fig:exp:mazevertical}}
    \end{subfigure}
    
    \begin{subfigure}[b]{0.49\linewidth}       \includegraphics[width=\figWidth,keepaspectratio]{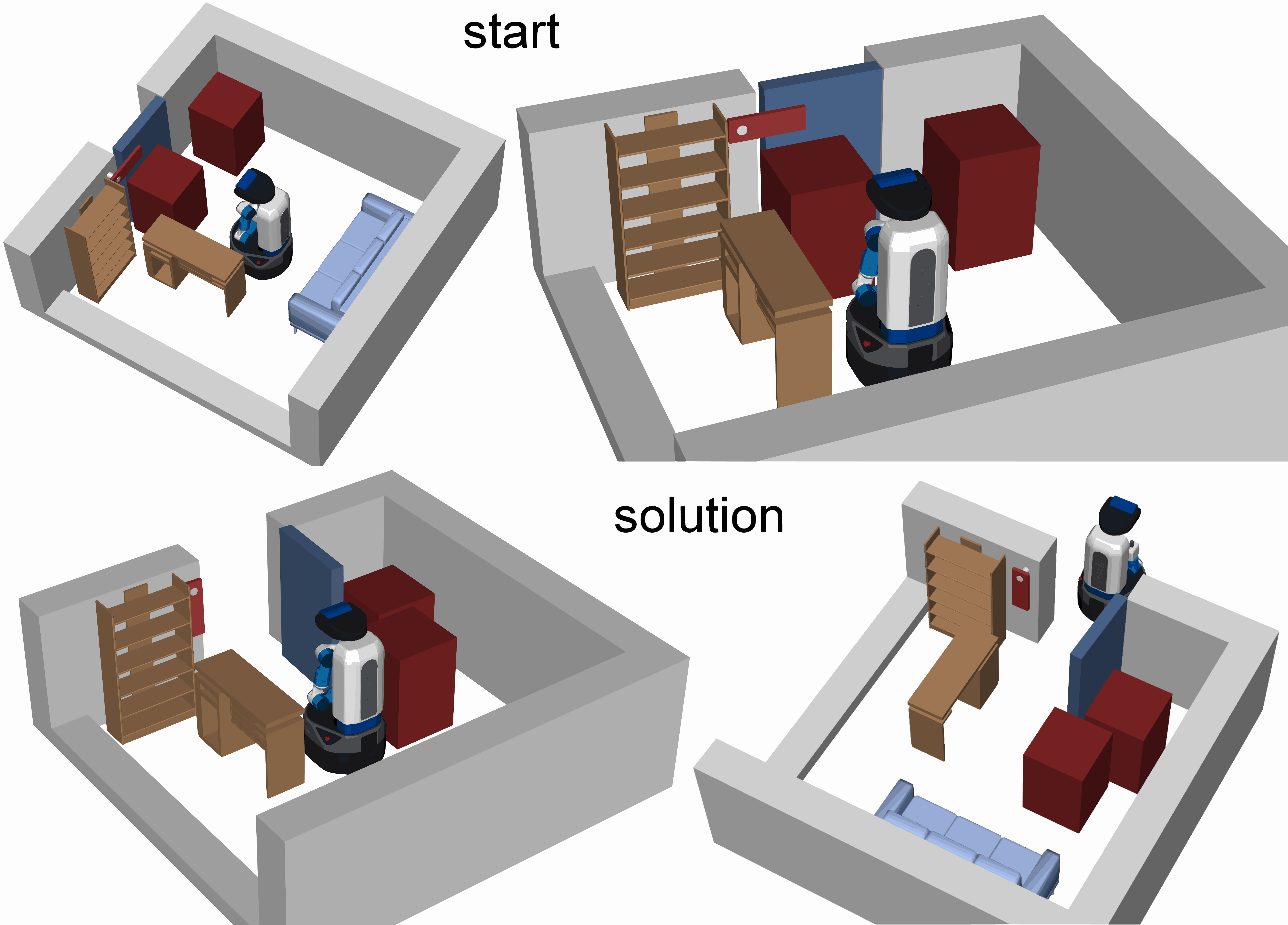}
    \caption{Escape Room 1\label{fig:exp:escaperoom}}
    \end{subfigure}
    \hfill
    \begin{subfigure}[b]{0.49\linewidth}       \includegraphics[width=\figWidth,keepaspectratio]{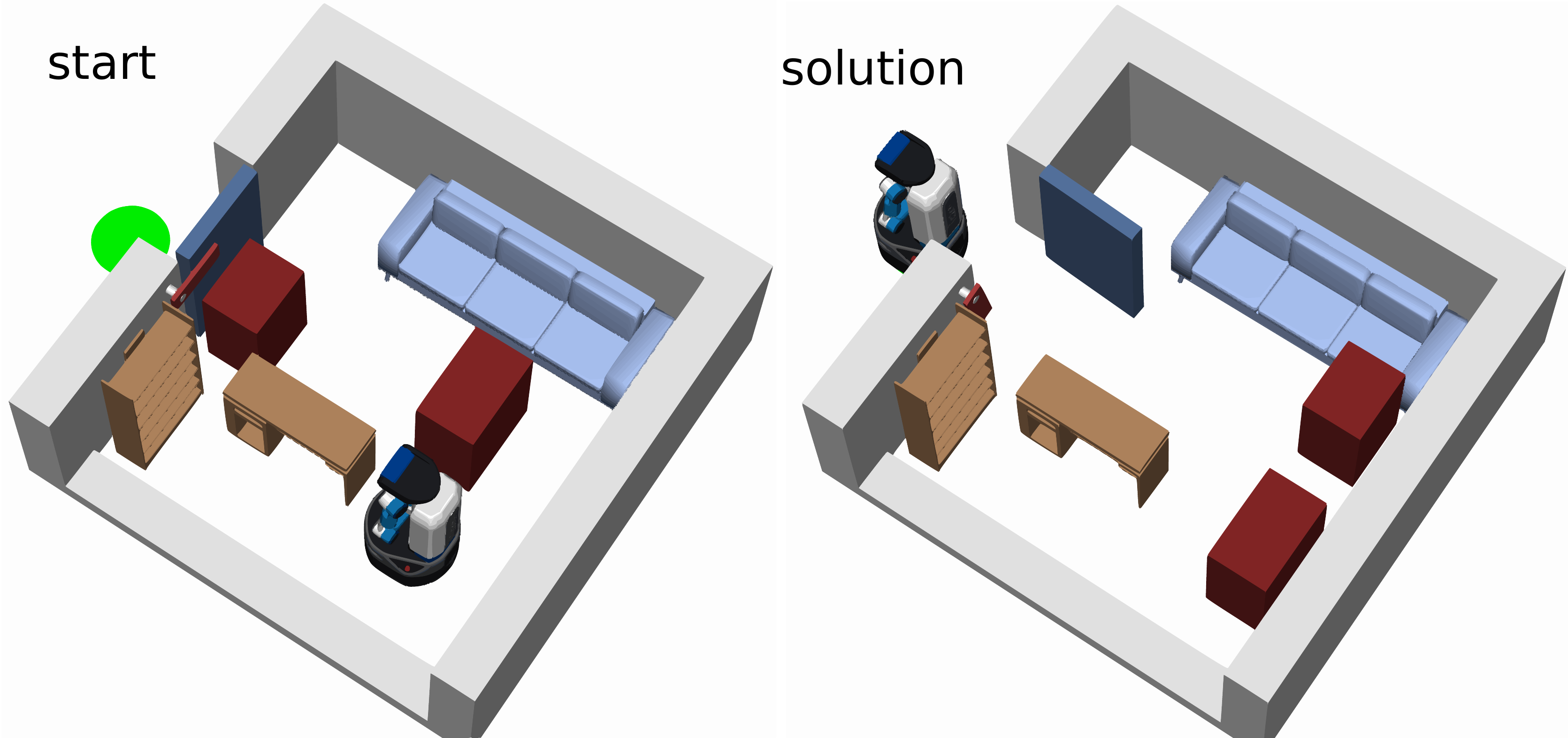}
    \caption{Escape Room 2\label{fig:exp:escaperoom2}}
    \end{subfigure}
    
    \caption{Experiments}
    \label{fig:my_label_exp}
\end{figure}

\def\subfigWidth{0.32\linewidth} 
\def\figWidth{\linewidth} 
\begin{figure*}
    \centering
    \begin{subfigure}{\subfigWidth}    
    \includegraphics[width=\figWidth,keepaspectratio]{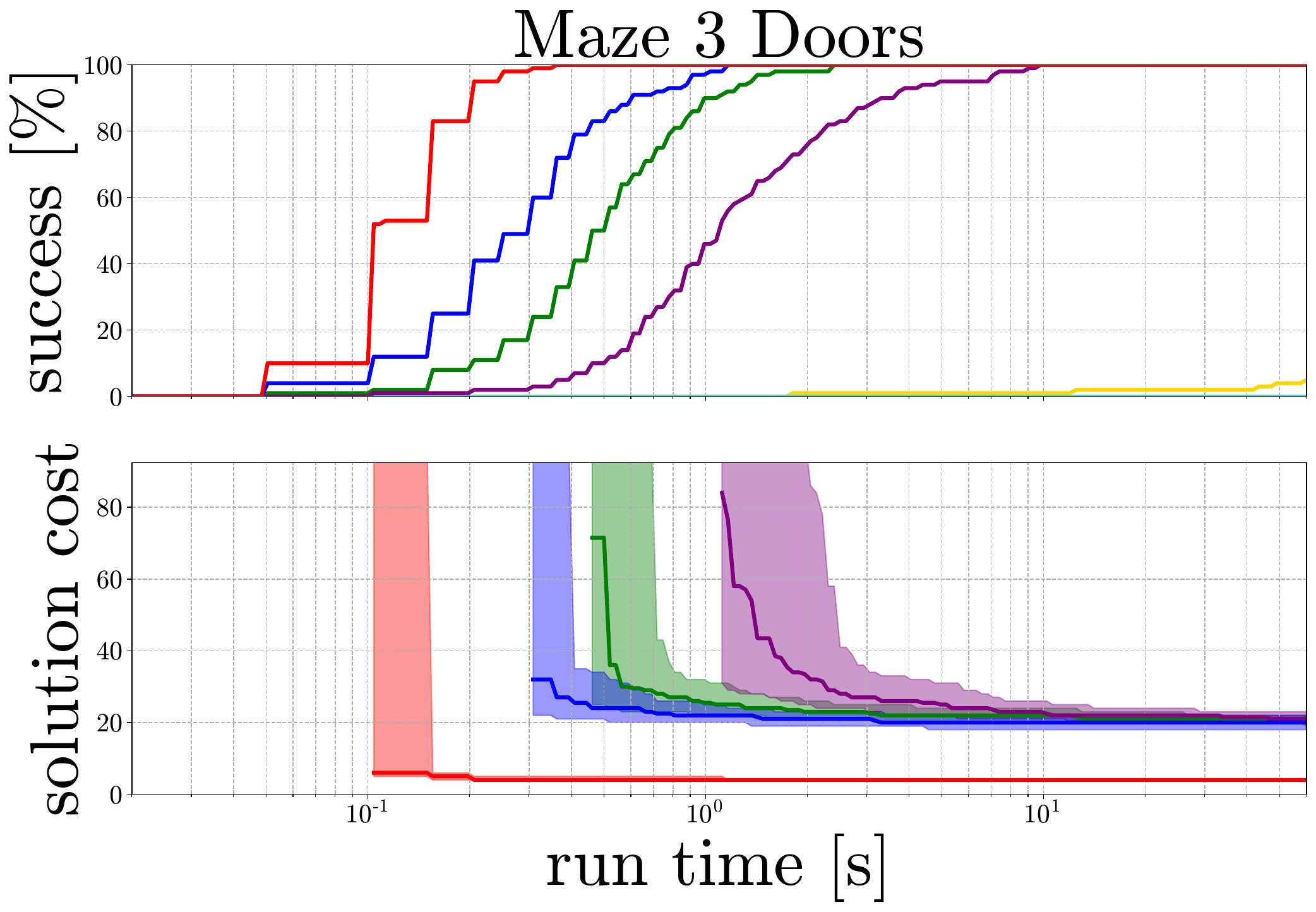}
    \caption{Maze 3 Doors}
    \end{subfigure}
    \begin{subfigure}{\subfigWidth}    
    \includegraphics[width=\figWidth,keepaspectratio]{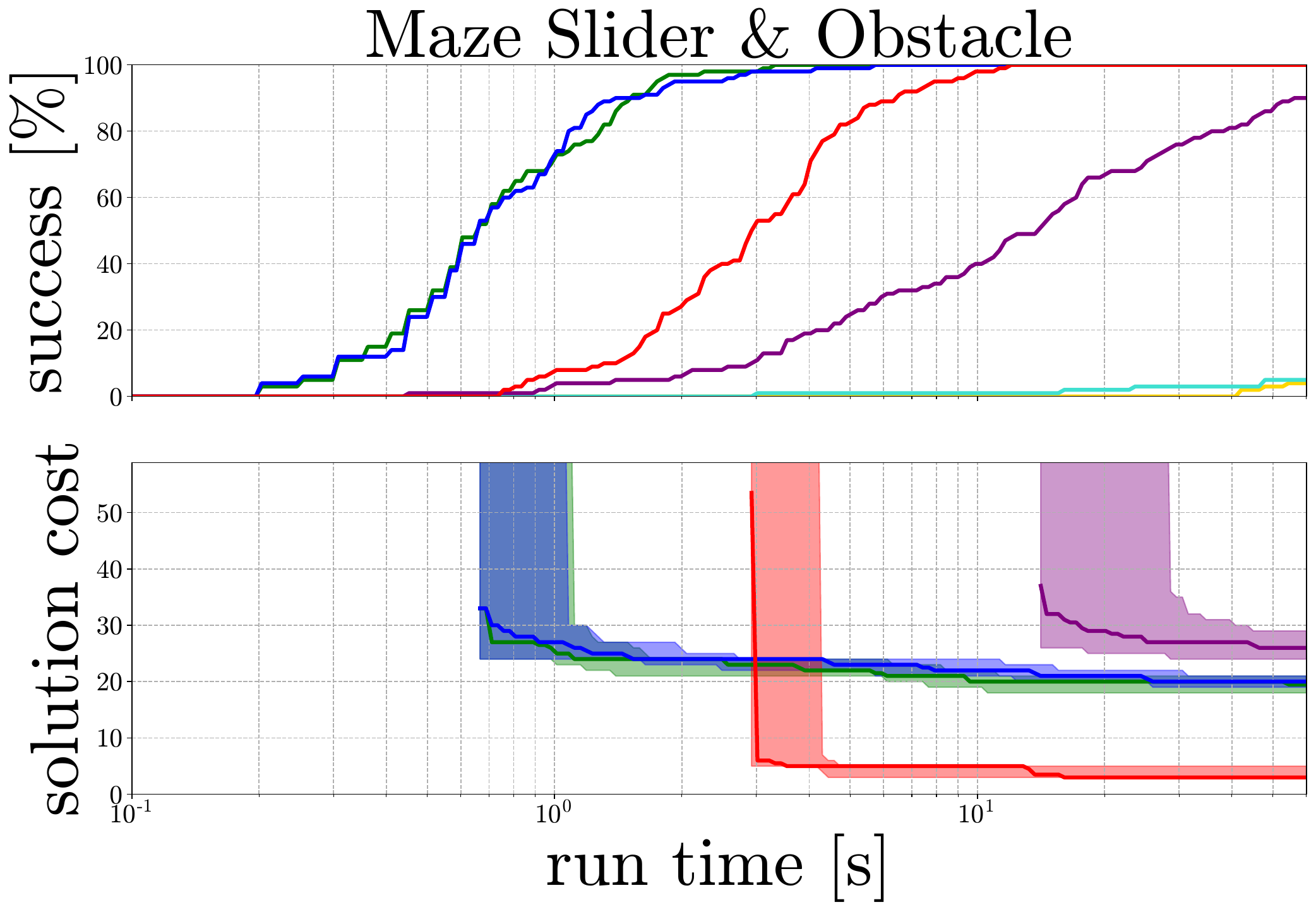}
    \caption{Maze Slider and Obstacle}
    \end{subfigure}
    \begin{subfigure}{\subfigWidth}       
    \includegraphics[width=\figWidth,keepaspectratio]{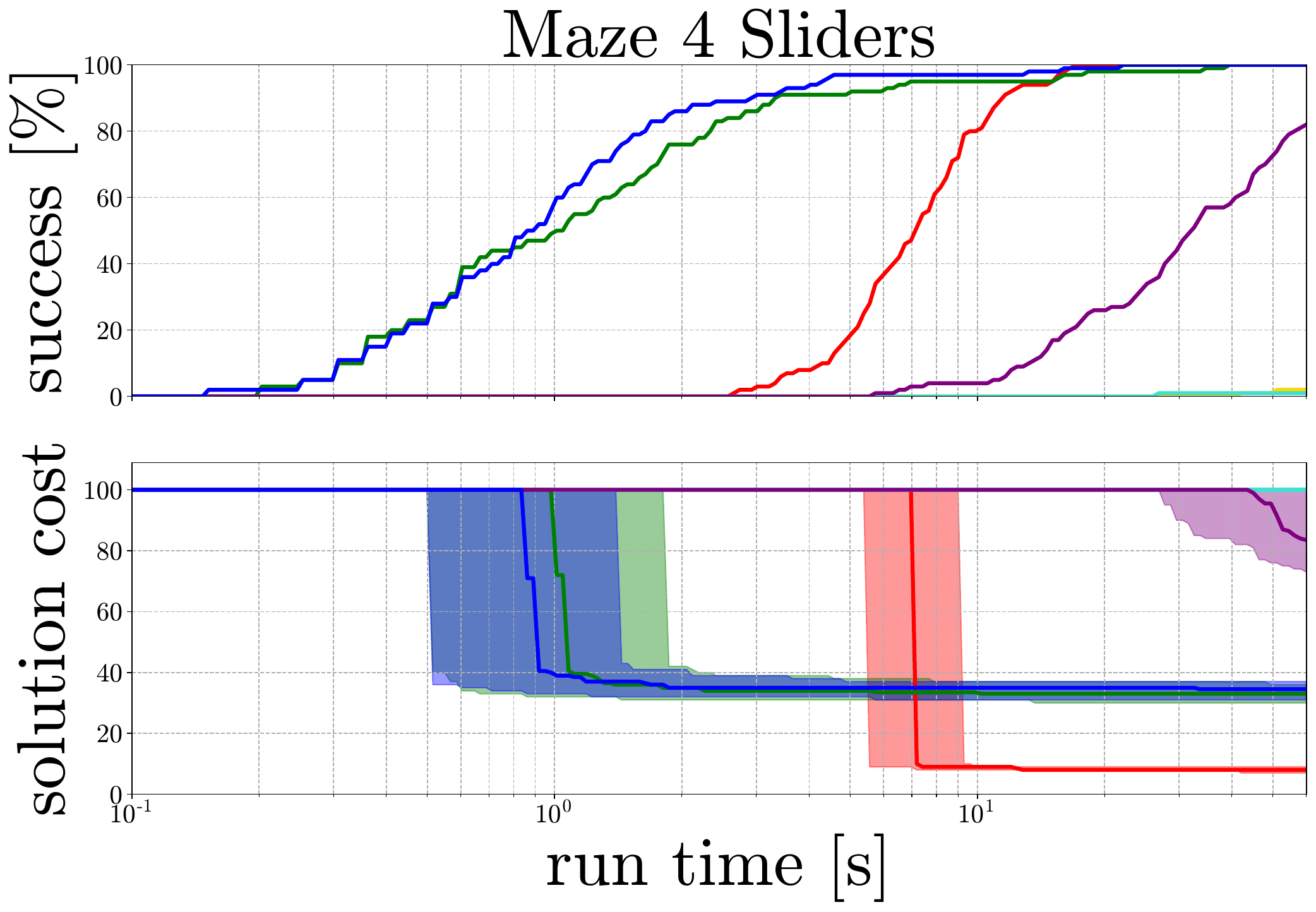}    
    \caption{Maze 4 Sliders}
    \end{subfigure}    
    \begin{subfigure}{\subfigWidth}       
    \includegraphics[width=\figWidth,keepaspectratio]{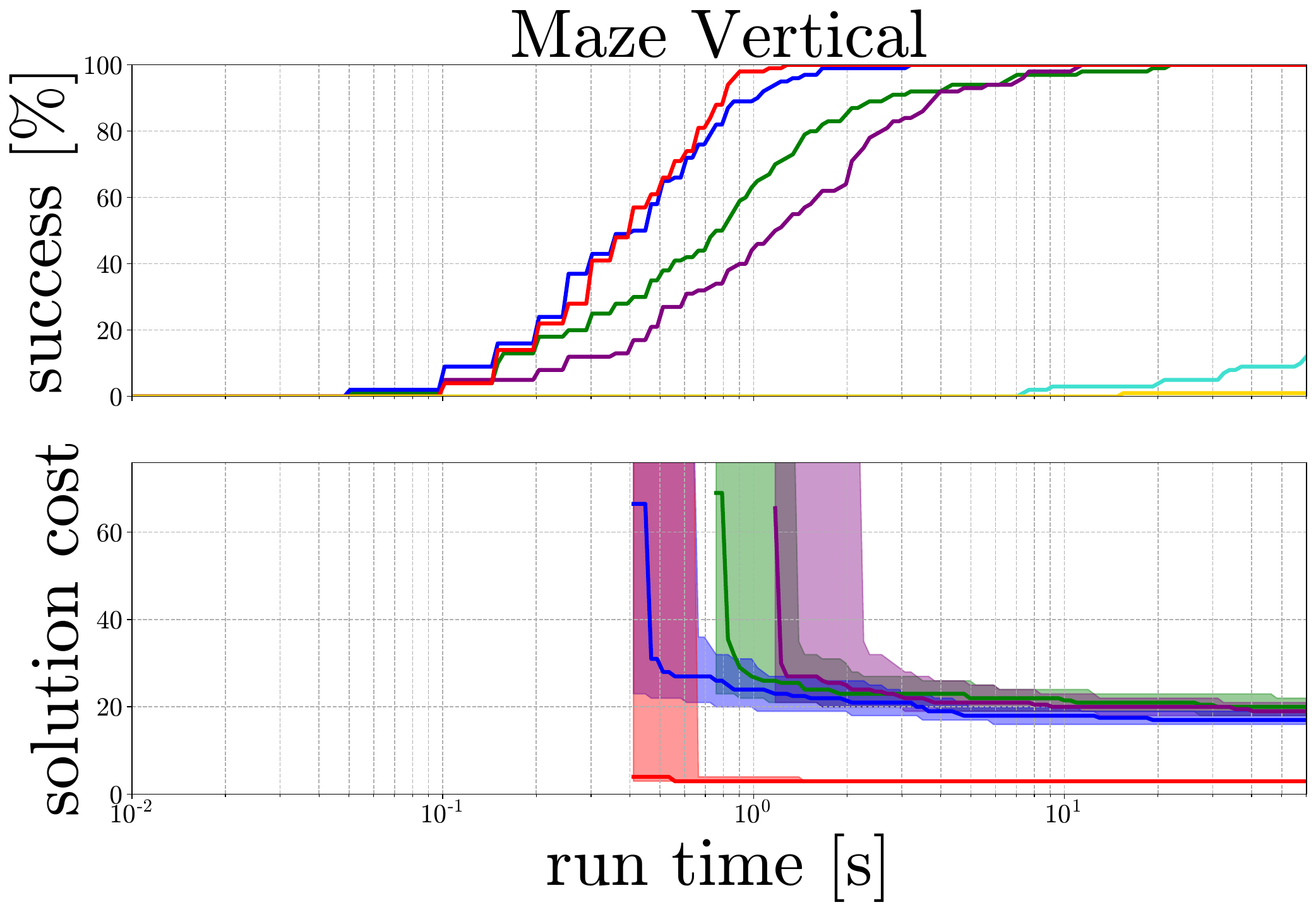}    
    \caption{Maze Vertical}
    \end{subfigure}    
    \begin{subfigure}{\subfigWidth}       
    \includegraphics[width=\figWidth,keepaspectratio]{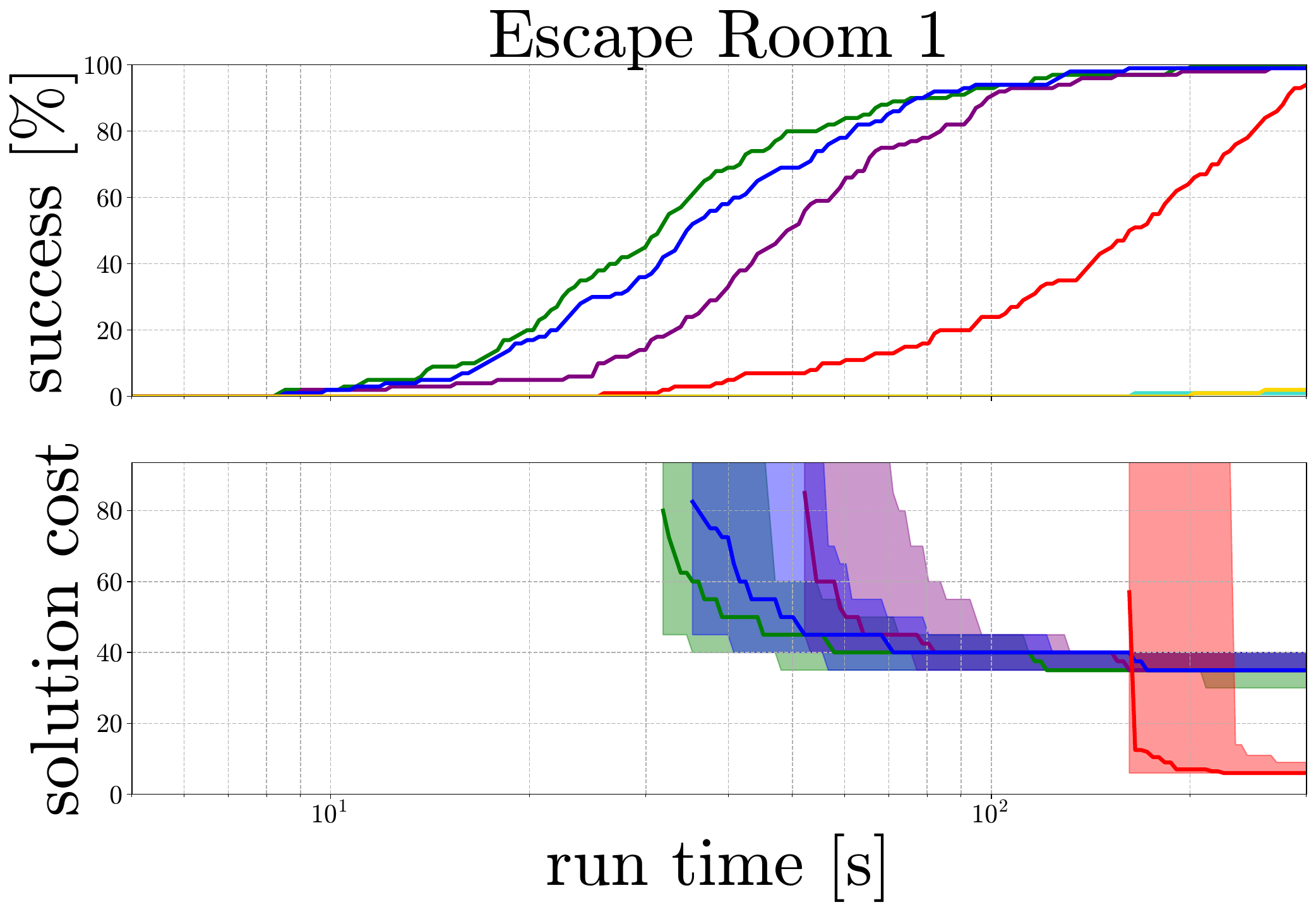} 
    \caption{Escape Room 1}
    \end{subfigure}
    \begin{subfigure}{\subfigWidth}       
    \includegraphics[width=\figWidth,keepaspectratio]{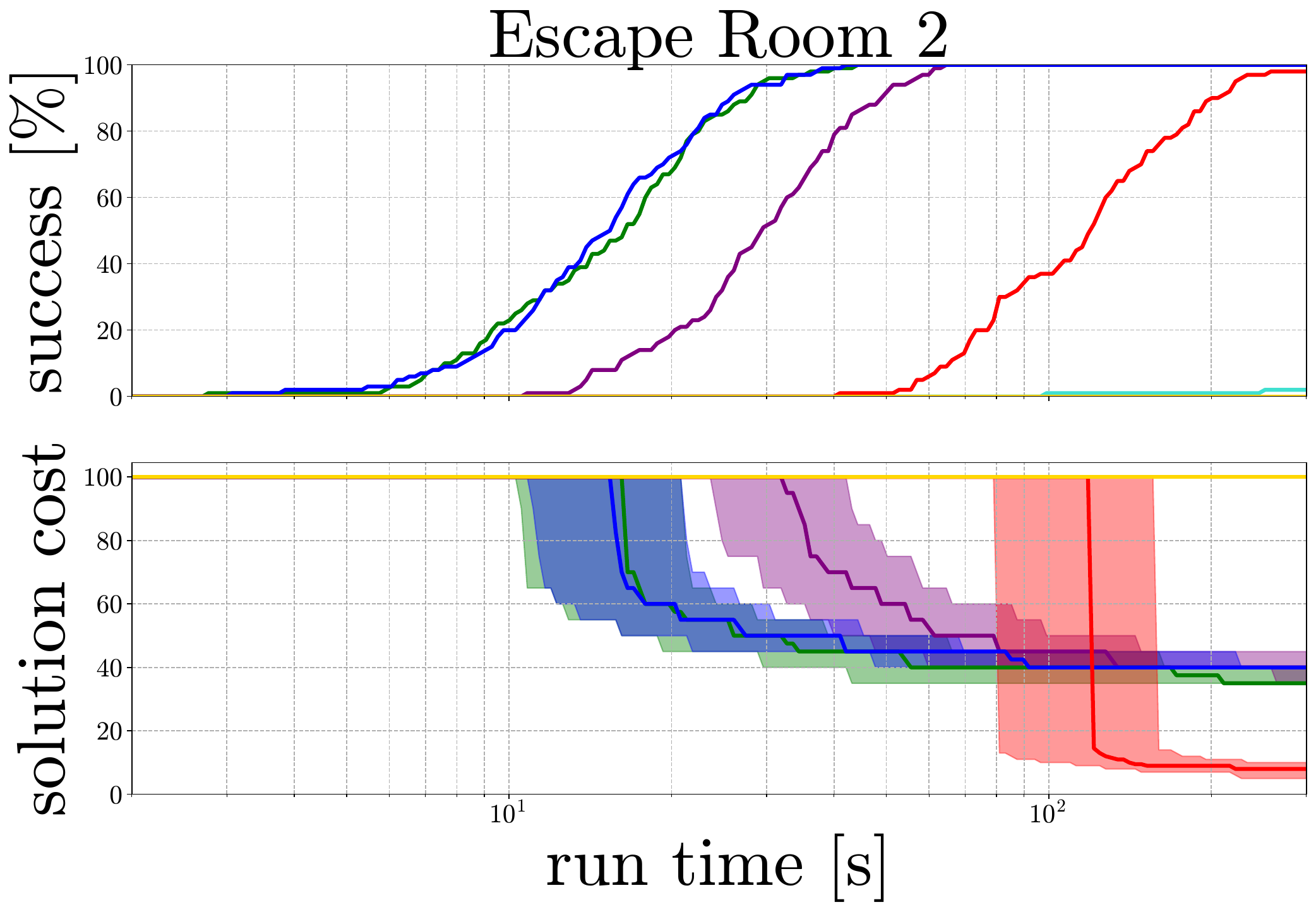} 
    \caption{Escape Room 2}
    \end{subfigure} 
    
    \caption{Experimental Results: \cbox{red} LA-RRT, \cbox{blue!50} BIT*,  \cbox{green} ABIT*, \cbox{orange!50} LBTRRT, \cbox{cyan} RRT*, and \cbox{violet!50} AIT*.}
    \label{fig:my_label}
\end{figure*}


In this section, we compare our algorithm and benchmark the success rate, cost, and solution time on the six following environments. For each environment, we define manually the factors for our factor state space.

\begin{itemize}
    \item{\textit{Maze 3 Doors} (\mf{fig:exp:maze3doors})}:
    The green cube has to move to a goal position (transparent green). Three doors with a hinge block its way, each rotatable from $-90^{\circ}$ to $90^{\circ}$. The best action cost to solve this puzzle is 4.
    
    \item{\textit{Maze Slider with Obstacle} (\mf{fig:exp:mazeslider})}:
    The green cube is blocked by a sliding door that is movable only on the x-axis and a freely movable cuboid on the x-y-axis inside the walls. This puzzle demonstrates that an object may need to be moved multiple times to reach the goal. The best action cost is 3.

    \item{\textit{Maze 4 Sliders} (\mf{fig:exp:maze4sliders})}:
    The cube has to move to the goal position through multiple sliding doors, and each door is completely movable along its rail. In the best case, each door would be moved only once so that the cube can directly be moved to the goal position, summing up to a total action cost of 5.
    
    \item{\textit{Maze Vertical} (\mf{fig:exp:mazevertical})}:
    Similar to the 3 Doors example, each door is rotatable between [$-90^{\circ}$,$90^{\circ}$] except the door on the top right which is rotatable between [$-90^{\circ}$,$0^{\circ}$]. The best cost, in this case, is 3.

    \item{\textit{Escape Room 1-2} (\mf{fig:exp:escaperoom,fig:exp:escaperoom2})}:
    In this scenario, a robot has to escape a room with fixed obstacles (desk, couch), and movable objects (cubes). The room has a lock and a door which can be rotated by [$0^{\circ}$, $90^{\circ}$] degrees. To account for the robot geometry, we add the robot as an additional movable object into the scene. For Escape Room 1 the best action cost is 4, consisting of moving only the cube in front of the door, unlocking the door, opening the door, and leaving the room. For Escape Room 2, it is 5 since both cubes need to be moved at least once.
    
\end{itemize}

\subsection{Implementation}
For the simulations, we use DART~\cite{lee2018dart} available through the Robowflex~\cite{kingston2021robowflex} library.
All evaluations are performed for 100 runs with a time limit of 100 seconds (except the escape room with a limit of 300 seconds) using the benchmarking tools of OMPL~\cite{sucan2012ompl, moll2015benchmarking}.
The goal of each scenario is to find the minimal number of actions, such that the robot needs the minimal number of pick-place actions to finish its task.
We benchmark \puzzlealgo against BIT*~\cite{gammell2020batch}, AIT*~\cite{strub2021ait}, ABIT*~\cite{strub2020abit}, RRT*~\cite{karaman2011sampling}, and LBTRRT*~\cite{salzman2016asymptotically} with default parameters using the factored state-space and the corresponding interpolate function. Because all those planners require an additive cost (\mf{sec:problemstatement}), we use the additive action cost for them. Only \puzzlealgo uses the non-additive action cost. For every planner, a goal space is defined for all goal indices, which leaves non-goal indices unspecified. The maximum number of states sampled in this goal space, $K$, is set to $K=10$. For visualization, we color all links actuated by goal-index joints in green and all other links in red. The desired goal configurations for the goal-index joints are shown in transparent green.
    
Since the result of those runs need to be sent to the robot, it is crucial to find the minimal number of actions. For this purpose, we only rely on optimal planners. Indeed, while other planners might find a feasible solution faster, this solution would not be valuable, since the action cost might lead to excessive pick-place actions which we would need to execute with the robot.

\subsection{Experimental Results}

\mf{fig:my_label} presents the benchmarking results of solving the problem in the object-only space. This is obtained from each problem environment, as shown in \mf{fig:my_label_exp}. The x-axis shows the runtime in seconds, and the y-axis shows the average success rate and solution cost.

In terms of success rate, LA-RRT achieves 100\% success in each environment except for \emph{Escape Room 1} (94\%) and \emph{Escape Room 2} (98\%). The RRT-based algorithms LBTRRT* and RRT* struggle to find a solution in the given time. BIT* and ABIT* achieve the best success rate in the fastest time, while AIT* performs a bit slower. In \emph{Maze 3 Doors} and \emph{Maze Vertical}, LA-RRT performs as fast as BIT* and ABIT*, but performs slower in the other \emph{Maze} environments and the \emph{Escape Room} scenarios. However, the quick success of BIT* and ABIT* comes at the price of a higher solution cost.


In every scenario, \puzzlealgo is able to exploit the non-additive action cost, and thereby reaches a significant lower solution cost than the next best planner. 
This situation is summarized in \mf{tab:averagesolution}, which shows the improvement in action cost compared to the next best planner. 
It can be seen that we can improve the action cost by $4.01$ to $6.58$ times, which is a significant improvement if we want to use those paths for manipulation tasks. 
As we discuss in \mf{sec:problemstatement}, other optimal planners cannot discriminate between paths with subsequent equivalent actions (\mf{fig:cost-explanation}), and therefore are not able to pick the correct equivalence class of paths, but not necessarily the true optimal solution.
 
\begin{center}
\centering\small
\begin{table}
\centering
\begin{tabular*}{\linewidth}
{@{\extracolsep{\fill}}p{0.27\linewidth}p{0.25\linewidth}P{0.10\linewidth}P{0.25\linewidth}@{}} 
 \hline
 Environment & Next Best Alg. & \puzzlealgo & Improvement \\ 
 \hline
 Maze 3 Doors & 19.74 (BIT*) & 3.00 & 6.58  \\ 
 \hline
 Maze Slider \& Obstacle & 19.2 (ABIT*) & 3.62 & 5.30 \\
 \hline
 Maze 4 Sliders & 33.39 (ABIT*) & 7.79 & 4.29 \\
 \hline
 Maze Vertical & 17.56 (BIT*) & 3.02 & 5.81 \\
 \hline
 Escape Room 1 & 35.40 (AIT*) & 7.11 & 4.98 \\ 
 \hline
 Escape Room 2 & 32.5 (ABIT*) & 8.12 & 4.01 \\  
 \hline
\end{tabular*}
	\caption{\label{tab:averagesolution}Average solution cost of LA-RRT and the next best algorithm for each environment.}
\end{table}
\end{center}

\section{Discussion and Conclusion}

We proposed to solve rearrangement puzzles using a new factored state space, which reflects the capabilities of the robot, without explicitly accounting for it. 
To properly exploit this state space, we developed the less-actions RRT (\puzzlealgo), which uses a path defragmentation method to optimize for a minimal number of actions, such that we minimize the number of pick-place actions our robot has to execute. 
In our evaluations, we showed that \puzzlealgo can consistently find lower number of actions compared to other state-of-the-art planners. 

While LA-RRT provides an admissible heuristic for the overall problem (i.e., a necessary condition to solve it), it can fail to produce a manipulatable solution. \mf{fig:my_label2} shows two scenarios, where the Fetch robot has to reach a green goal configuration, while opening doors and removing a red cube. In~\mf{fig:exp:limitcase1}, the scenario is solvable by LA-RRT, but the resulting solution is not executable by the Fetch robot. Such situations could be overcome by backtracking on the manipulation level. In~\mf{fig:exp:limitcase2}, the red cube blocks the door and makes the scenario infeasible. This is a well-known limitation of sampling-based planners, and could be addressed by improving infeasibility checking~\cite{li2021learning}. Despite those limitations, having a good, informed admissible heuristic is important to efficiently solve difficult rearrangement puzzles.

In summary, we successfully showed that LA-RRT can be used to solve rearrangement puzzles so that final paths with low action costs can be found, making them executable by a robot in a realistic time. We believe this to be an important step towards an integrated framework for efficient planning of high-dimensional rearrangement puzzles.

\def\figWidth{0.32\linewidth} 

\def\subfigWidth{\linewidth} 
\def\figWidth{0.9\linewidth} 
\begin{figure}   
    
    \centering
    \begin{subfigure}{\subfigWidth}   
    
    \centering
    \includegraphics[width=\figWidth,keepaspectratio]{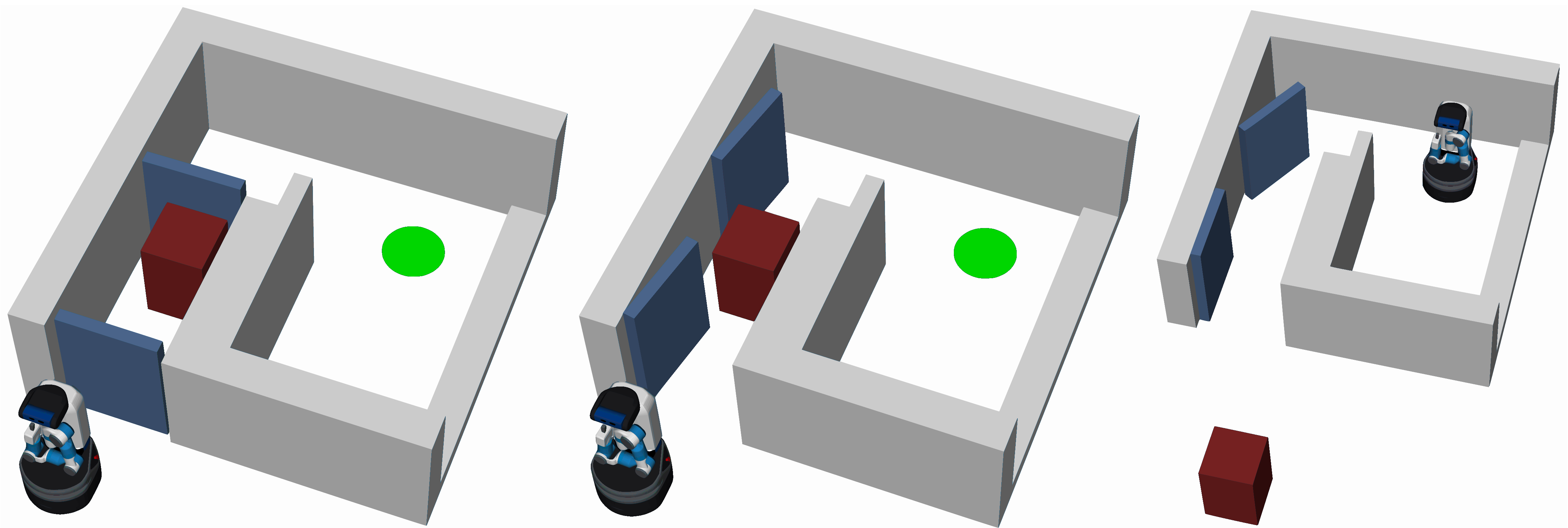}
    \caption{Limitation 1: The cube in front of the interior door blocks the robot.\label{fig:exp:limitcase1}}
    \end{subfigure}
    
    \begin{subfigure}{\subfigWidth}
    
    \centering
    \includegraphics[width=\figWidth,keepaspectratio]{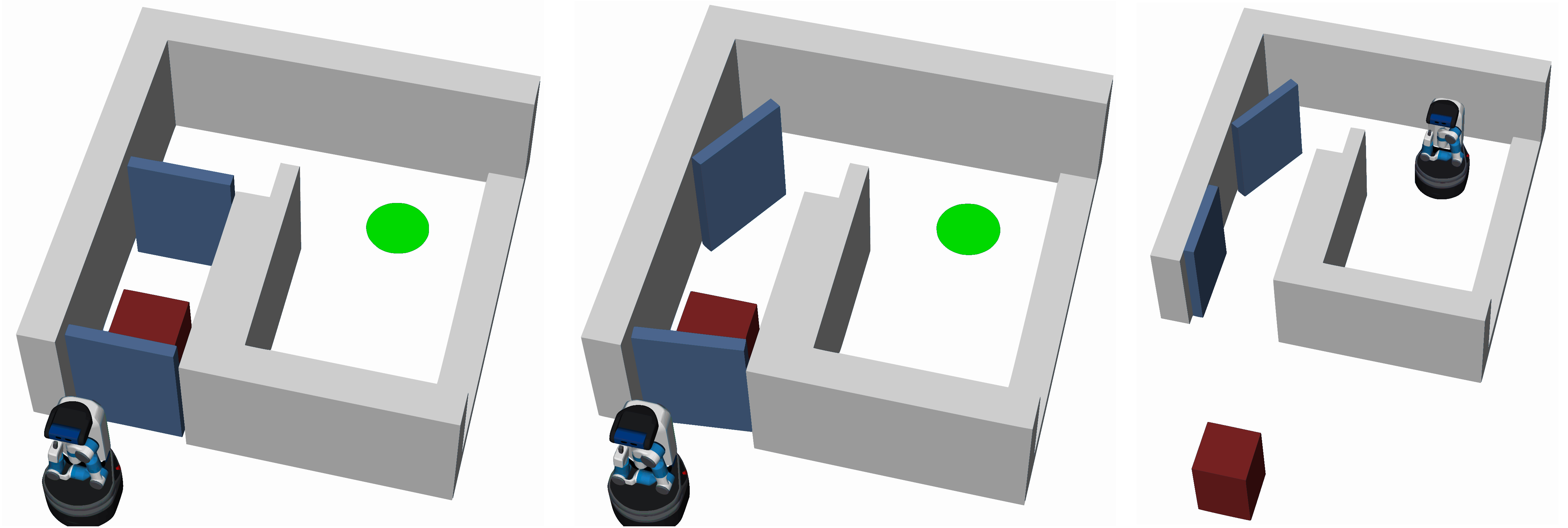}
    \caption{Limitation 2: The cube behind the front door blocks the door.\label{fig:exp:limitcase2}}
    \end{subfigure}
    
    \caption{Environments solvable by \puzzlealgo but are not feasible.}

    \label{fig:my_label2}
\end{figure}

    
    

\bibliographystyle{IEEEtranS}
{
\balance
\small
\bibliography{IEEEabrv, bib/general}
}

\end{document}